\documentclass[letterpaper]{article}
\usepackage{aaai2026}
\usepackage{times}
\usepackage{helvet}
\usepackage{courier}
\usepackage[hyphens]{url}
\usepackage{graphicx}
\usepackage{natbib}
\usepackage{caption}
\usepackage{algorithm}
\usepackage{algorithmic}
\usepackage{amsfonts}
\usepackage{mathrsfs}
\usepackage{booktabs}
\usepackage{multirow}
\usepackage{amsmath}

\usepackage{newfloat}
\usepackage{listings}
\DeclareCaptionStyle{ruled}{labelfont=normalfont,labelsep=colon,strut=off}
\floatstyle{ruled}
\newfloat{listing}{tb}{lst}{}
\floatname{listing}{Listing}

\title{Graph Negative Feedback Bias Correction Framework for Adaptive Heterophily Modeling}

\author{
    Jiaqi Lv\textsuperscript{1},
    Qingfeng Du\textsuperscript{1,*},
    Yu Zhang\textsuperscript{1},
    Yongqi Han\textsuperscript{1},
    Sheng Li\textsuperscript{1}
}

\affiliations{
    \textsuperscript{1}School of Computer Science and Technology, Tongji University, Shanghai, China\\
}

\usepackage{bibentry}

\begin{document}

\maketitle
    
\begin{abstract}
Graph Neural Networks (GNNs) have emerged as a powerful framework for processing graph-structured data. However, conventional GNNs and their variants are inherently limited by the homophily assumption, leading to degradation in performance on heterophilic graphs. Although substantial efforts have been made to mitigate this issue, they remain constrained by the message-passing paradigm, which is inherently rooted in homophily. In this paper, a detailed analysis of how the underlying label autocorrelation of the homophily assumption introduces bias into GNNs is presented. We innovatively leverage a negative feedback mechanism to correct the bias and propose Graph Negative Feedback Bias Correction (GNFBC), a simple yet effective framework that is independent of any specific aggregation strategy. Specifically, we introduce a negative feedback loss that penalizes the sensitivity of predictions to label autocorrelation. Furthermore, we incorporate the output of graph-agnostic models as a feedback term, leveraging independent node feature information to counteract correlation-induced bias guided by Dirichlet energy. GNFBC can be seamlessly integrated into existing GNN architectures, improving overall performance with comparable computational and memory overhead.
\end{abstract}

\section{Introduction}

Graph Neural Networks (GNNs) have emerged as a powerful framework for processing graph-structured data, with applications spanning recommendation systems \cite{recommendation_systems}, e-Commerce platforms \cite{e-Commerce_platforms}, and social networks \cite{social_networks}. A key strength of GNNs lies in their ability to enhance node features by incorporating topological information through message propagation during graph convolution. However, the effectiveness of this approach is limited by the assumption of homophily, where similar nodes are more likely to be connected. In graphs exhibiting low homophily, or heterophily, graph convolution can lead to performance degradation, sometimes underperforming convolution-free models like Multi-Layer Perceptrons (MLPs).

Recently, numerous methods have been proposed to address the challenges posed by heterophily in GNNs. Most of these methods are implemented by modifying the graph structure or designing personalized neighbor aggregation strategies. LHS \cite{LHS} learns latent homophilic structures through self-expressive techniques and dual-view contrastive learning to mitigate structural threats outside of distribution. H2GCN \cite{H2GCN} enhances node features in heterophily graphs by separating ego- and neighbor-embedding and integrating higher-order neighbor information. Other methods, such as H2-FDetector \cite{H2-FDetector} and GBK-GNN \cite{GBK-GNN}, aggregate information separately from homophilic and heterophilic connections. However, these methods remain constrained by the homophily assumption, as the fundamental aggregation strategy of GNNs has not been altered. This example raises a crucial question. \textit{How can we design a general modeling framework for graphs with varying degrees of heterophily, without being limited to specific aggregation strategies?}

\begin{figure}[htbp]
    \centering
    \includegraphics[width=0.95\linewidth]{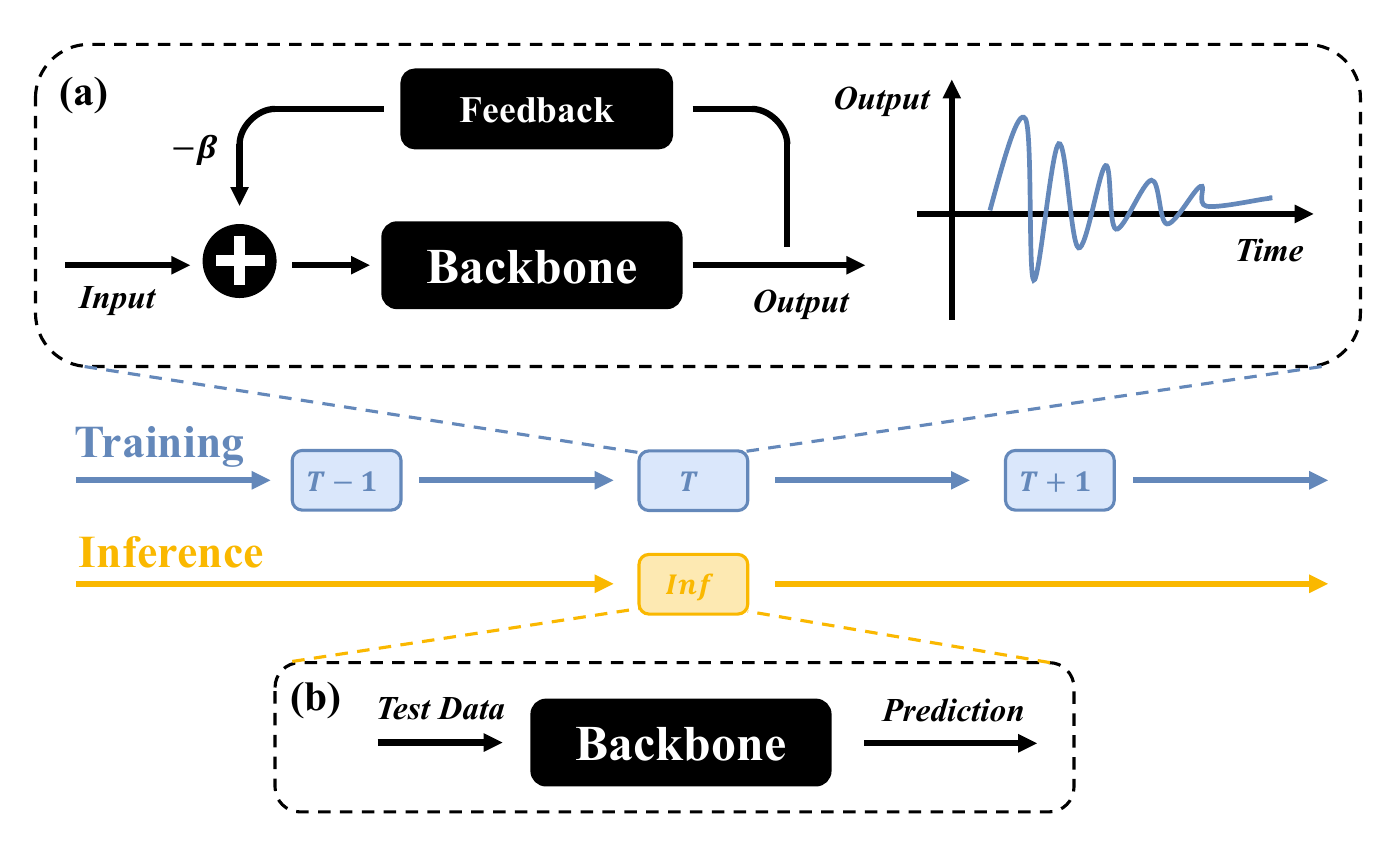}
    \caption{Schematic diagram of the negative feedback system and its adjustment process toward a new steady state. (a) denotes the training process, and (b) denotes the inference process.}
    \label{fig:introduction}
\end{figure}

The essence of the homophily assumption is that it introduces label autocorrelation in GNNs and autocorrelation inevitably introduces bias during the learning process. From an information-theoretic perspective, the node classification task with cross-entropy loss aims to maximize the mutual information between the graph and the provided labels. If the labels are autocorrelated, the graph structure may capture redundant information between labels rather than genuine topological dependencies. Furthermore, in a fixed topology, the sharing of node features can be obscured by label autocorrelation. If the autocorrelation of the labels is stronger than the independent information of node features, the model may underestimate the role of node features. In contrast, graph-agnostic models, such as MLPs, do not introduce label autocorrelation. Therefore, the output of graph-agnostic models can be used to apply negative feedback correction to GNNs. The negative feedback mechanism \cite{negative_feedback} can suppress the noise in the input signal, stabilizing the system in a set state. As shown in Figure \ref{fig:introduction}, during the training process, the feedback model is used to correct the bias of the backbone model, ultimately bringing the signal to a steady state. In the inference process, only the backbone model is involved, introducing no additional computational overhead. For each GNN model, an equivalent graph-agnostic model can be obtained by removing the aggregation step. Therefore, the parameters of the two models can be shared to further reduce the space overhead.

Thus, we propose Graph Negative Feedback Bias Correction (GNFBC), a straightforward yet effective framework to correct biases due to incomplete modeling of label autocorrelation. First, we introduce a negative feedback loss to penalize the sensitivity of predictions to label autocorrelation. This makes the model focus more on learning topological dependencies rather than on the degree of heterophily of the graph itself. Moreover, we use the output of graph-agnostic models as a feedback term, leveraging independent information from node features to compensate for correlation biases. We utilize Dirichlet energy to model the degree of autocorrelation of nodes, determining the extent of correction needed during the iterative process. GNFBC can be seamlessly integrated into almost any GNN architecture, improving overall model performance with comparable computational and space overhead. In summary, GNFBC distinguishes itself from previous work through some key features:

\begin{itemize}
    \item GNFBC corrects biases caused by incomplete modeling of label autocorrelation, improving overall model performance, and enhancing the generalization of GNNs across graphs with varying degrees of heterophily.
    \item As a universal framework, GNFBC can be seamlessly integrated into almost any GNN architecture. Meanwhile, it has comparable computational and memory overhead.
    \item We perform a comprehensive evaluation of GNFBC on graphs with varying degrees of heterogeneity. Extensive experimental results demonstrate that GNFBC exhibits exceptional competitiveness.
\end{itemize}

\section{Related Work}

Most graph neural networks (GNNs) rely on the homophily assumption, which posits that connected nodes tend to share similar features or labels. This assumption underpins the design of neighborhood aggregation schemes that propagate and mix information from adjacent nodes. Although effective in many applications, such methods often suffer performance degradation when applied to heterogeneous graphs, where neighboring nodes are more likely to have different attributes or belong to different classes. Recent works have highlighted this limitation and proposed a variety of approaches to address it. One line of research designs new aggregation mechanisms that decouple feature smoothing from label prediction, such as Geom-GCN \cite{Geom-GCN} and H2GCN \cite{H2GCN}, which aim to preserve node-specific information while selectively aggregating useful neighbor signals. Another line explores adaptive weighting of neighbors based on feature or label similarity \cite{BM-GCN}, using techniques such as attention \cite{HAN}, learnable filters \cite{learnable_filters}, or neighbor selection \cite{neighbor_selection}. In addition, some studies investigate alternative propagation schemes or model architectures that are more robust to heterophily, including graph transformers \cite{graph_transformer} and higher-order neighborhood modeling \cite{Mixhop}. Despite these advances, effectively capturing informative patterns in highly heterogeneous graphs while avoiding overfitting and over-smoothing remains an open challenge.

\section{Preliminaries}

A graph is indicated as $\mathcal{G} = \left \{ \mathcal{V},\mathcal{E},\mathcal{X},\mathcal{Y} \right \}$, where $\mathcal{V}$ is the node set and $\mathcal{E}$ is the edge set. $x_i \in \mathcal{X}$ represents a \textit{d}-dimension feature vector of node $v_i$, where $x_i \in \mathbb{R}^d$; $\mathcal{Y}$ is the set of labels for each node. For an undirected graph, the structure of the graph can be represented by an adjacency matrix $\mathcal{A}\in \mathbb{R}^{N\times N}$, where $N$ is the number of nodes. The graph node classification aims to predict the node labels $\mathcal{Y}$ utilizing node features $\mathcal{X}$ and topological information.

\subsection{Homophily and Heterophily}

Homophily and heterophily describe the connectivity patterns in the graphs. Homophily refers to the tendency of nodes with similar attributes (e.g., labels or features) to form edges, resulting in tightly knit communities. The most widely adopted definition of homophily is edge homophily, which is defined as the average agreement between pairs of adjacent labels.

\begin{equation}
    \mathcal{H}\left ( \mathcal{G}  \right ) = \frac{1}{\left | \mathcal{E}  \right |    } \sum_{\left ( u,v \right ) \in \mathcal{E} }^{} \textbf{1}\left ( y_u = y_v \right )
\end{equation}

Conversely, heterophily describes nodes connecting to dissimilar nodes, common in networks like fraud detection or protein interaction graphs. However, a notable defect of label-related homophily definitions is that they may be ill-defined when the underlying graph is label-scarce, which is often encountered in practice.

\subsection{Negative Feedback}

Negative feedback is a mechanism that reduces or suppresses changes in the input signal, thereby stabilizing the system or maintaining it at a set state \cite{Process_control}. The core function of negative feedback is to detect deviations in the system's output and automatically adjust the input to minimize these deviations, ensuring the system's balance or stable operation. If a system is represented by $y=H(x)$, negative feedback can be seen as adding the following quantity to the input to counteract changes in the output.

\begin{equation}
    x_{neg} = -k\Delta y, k>0
\end{equation}

\subsection{Graph-aware and Graph-agnostic Models}

A network that incorporates feature aggregation based on graph structure is termed a graph-aware model, e.g., GCN, SGC. In contrast, graph-agnostic models, such as MLPs, process node features independently, ignoring connectivity. A graph-aware model is always coupled with a graph-agnostic model. When the aggregation step is removed, a graph-aware model reduces to its graph-agnostic equivalent, e.g. GCN is coupled with MLP with 2 layers as shown below.

\begin{equation}
    GCN:Softmax\left ( \tilde{A}ReLU\left ( \tilde{A}XW_0 \right ) W_1 \right )
\end{equation}

\begin{equation}
    MLP: Softmax\left ( ReLU\left ( XW_0 \right ) W_1 \right ) 
\end{equation}

\section{Theoretical Analysis}

The autocorrelation of neighbor labels is a fundamental characteristic of graph data. Vanilla GNNs explicitly model this correlation through the homogeneity assumption, while networks designed for heterogeneous graphs redefine this correlation by modifying the aggregation strategy. The aggregation strategy needs to be adjusted based on varying degrees of heterogeneity, which limits the generalization of the model. From an information-theoretic perspective, the node classification task with cross-entropy loss aims to maximize the mutual information between the graph and the provided labels, denoted $I\left ( \mathcal{G},\mathcal{Y}_l   \right )$. If we consider graph $\mathcal{G}$ as a joint distribution of node features $\mathcal{X}$ and edges $\mathcal{E}$, $I\left ( \mathcal{G};\mathcal{Y}_l   \right )$ can be decomposed into a sum of conditional mutual information.

\begin{equation}
    I\left ( \mathcal{G};\mathcal{Y}_l   \right ) = I\left ( \mathcal{X},\mathcal{E};\mathcal{Y}_l    \right ) = I\left ( \mathcal{E}; \mathcal{Y}_l \right ) + I\left ( \mathcal{X};\mathcal{Y}_l\mid \mathcal{E}    \right )   
\end{equation}

The first term represents the information encoded in the graph structure, while the second term represents information from the node features. If $\mathcal{Y}_l$ exhibits autocorrelation, $I\left ( \mathcal{G};\mathcal{Y}_l   \right )$ would overestimate the explanatory power of the graph structure and features of the labels, as part of the mutual information originates from the dependencies between the labels rather than the independent contribution of $\mathcal{G}$. This misalignment introduces bias into the learning objective of the node classification task. For simplicity, we only model the bias caused by autocorrelation of first-order neighbor node labels.

\begin{equation}
    \epsilon_i = \hat{\mathcal{Y}_i } + {\textstyle \sum_{j\in\mathcal{N}\left ( i \right )  }^{}} \rho_{ij}\left ( \mathcal{Y}_j-\hat{\mathcal{Y}_j }   \right )
\end{equation}

\begin{equation}
    Bias \approx \sum_{i=1}^{N} \frac{1}{2 \sigma^2 } \left ( \mathcal{Y}_i - \hat{\mathcal{Y}_i }   \right )^2 -\sum_{i=1}^{N} \frac{1}{2\sigma^2\left ( 1-\rho_i^2 \right ) } \left ( \mathcal{Y}_i - \epsilon_i  \right ) ^2 
\end{equation}

$\hat{\mathcal{Y}_i }$ indicates the prediction of node $i$, $\rho_{ij}$ denotes the partial correlation between $\mathcal{Y}_i$ and $\mathcal{Y}_j$, $\rho_i^2 =  {\textstyle \sum_{j\in\mathcal{N}\left ( i \right )  }^{}} \rho_{ij}^2$. The presence of label autocorrelation $\rho_{ij}$ causes the loss to be biased against the NLL of the real data. If negative feedback is introduced during aggregation and continuously applied for correction, it can effectively reduce $\rho_i^2$, thus mitigating the bias. Let a learnable negative feedback coefficient $\gamma_{ij}$ be introduced, and the adjusted correlation coefficient is as follows:

\begin{equation}
    \rho ^{'}_{ij} = \rho_{ij}\left ( 1-\gamma_{ij} \right ) 
\end{equation}

The objective of model training should explicitly consider how to design a negative feedback mechanism through $\gamma_{ij}(\theta)$ to reduce $\rho_i^2$, improving the unbiasedness and robustness of node classification. In the next section, we will introduce how to design the negative feedback mechanism.

\section{Methodology}

\begin{figure*}[htbp]
    \centering
    \includegraphics[width=0.95\textwidth]{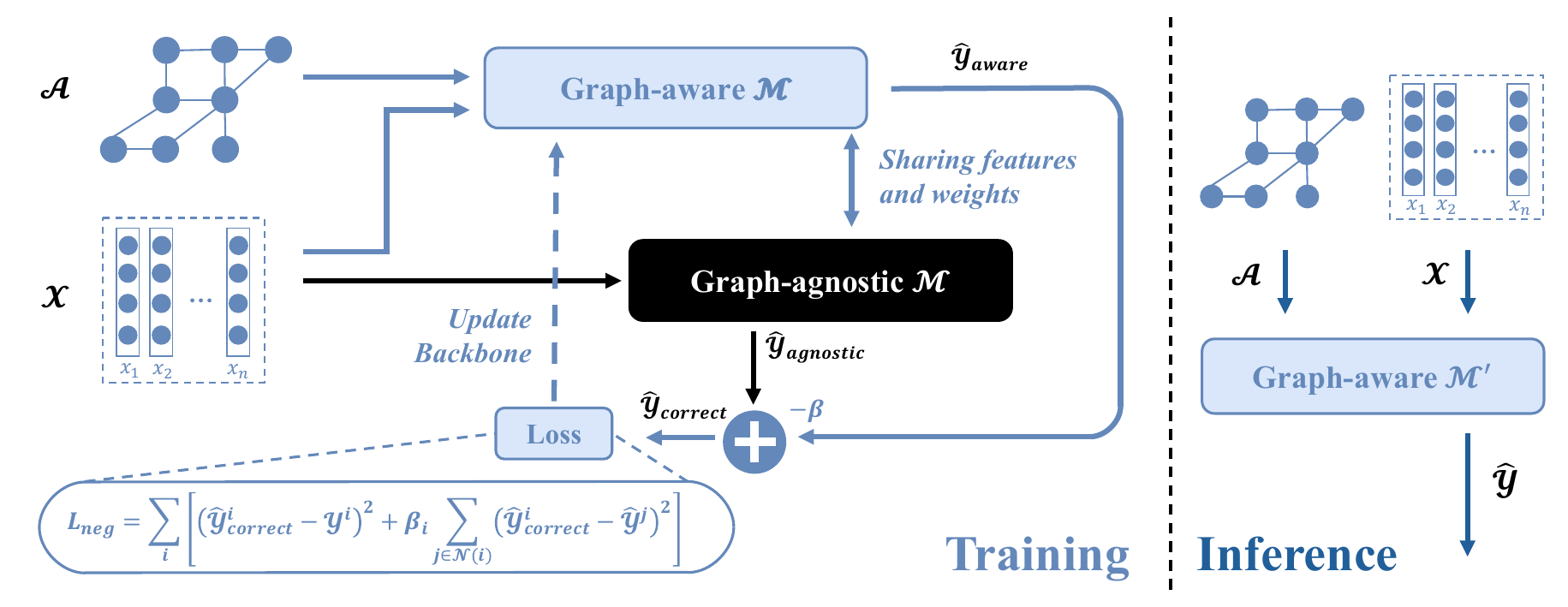}
    \caption{GNFBC employs a graph-agnostic model to perform negative feedback correction during training. The graph-aware model and the graph-agnostic model share features and weights. During the inference stage, GNFBC follows the standard inference process.}
    \label{fig:GNFBC}
\end{figure*}

In this section, we present the proposed GNFBC framework in detail. As shown in Figure \ref{fig:GNFBC}, we use the graph-agnostic model to perform negative feedback correction during the training process. In the inference stage, the process remains the same as in the standard inference. Next, we will provide a detailed introduction to each module of the negative feedback correction.

\subsection{Negative Feedback Loss}

The node classification task typically optimizes the matching between $\hat{\mathcal{Y}}$ and $\mathcal{Y}$ using cross-entropy or mean squared error.

\begin{equation}
    L = \mathbb{E}\left [ \left ( \hat{\mathcal{Y}}-\mathcal{Y}    \right )^2  \right ]  
\end{equation}

We can incorporate a penalty term into the loss function to discourage predictions $\hat{\mathcal{Y}}$ from over-relying on the autocorrelation of labels. Specifically, assuming that the autocorrelation propagates to $\mathcal{Y}_i$ through the neighbor labels $\mathcal{Y}_j$, we define the negative feedback term as follows:

\begin{equation}
    L_{neg}=\sum_{i}^{} \left [ \left ( \hat{\mathcal{Y}_i } -\mathcal{Y}_i  \right )^2 + \beta_i \sum_{j\in\mathcal{N}\left ( i \right )  }^{} \left ( \hat{\mathcal{Y}_i } -\hat{\mathcal{Y}_j}  \right )^2  \right ] 
\end{equation}

The first term is the standard error and the second term is the negative feedback term, where $\beta_i>0$ controls the degree of the autocorrelation penalty. Since each node exhibits a different level of autocorrelation, we apply different degrees of correction to each node accordingly. 
The specific computation of $\beta_i$ will be introduced in the final subsection of this chapter.

\subsection{Graph-Agnostic Model Feedback Correction}

Graph-aware models exploit the structural information of neighboring nodes in the graph.

\begin{equation}
    \hat{\mathcal{Y}}_i^{aware} = f_{aware}\left ( x_i, \mathcal{A} ;\theta  \right ) 
\end{equation}

In contrast, graph-agnostic models disregard the structural information of the graph and rely exclusively on the intrinsic features of individual nodes. Notably, a graph-aware model reduces to its graph-agnostic counterpart if the neighbor feature aggregation mechanism is omitted or the adjacency matrix is replaced by the identity matrix. Meanwhile, the parameters are shared between the two models.

\begin{equation}
    \hat{\mathcal{Y}}_i^{agnostic} = f_{agnostic}\left ( x_i;\theta  \right ) = f_{aware}\left ( x_i,\mathcal{A}=I;\theta  \right ) 
\end{equation}

We define the ground truth as $\mathcal{Y}$, and aim to approximate $\mathcal{Y}$ as closely as possible. The prediction of a graph-aware model can be further expressed as:

\begin{equation}
    \hat{\mathcal{Y}}_i^{aware} = f_{agnostic}\left ( x_i \right ) + f_{homo}\left ( \left \{ x_j:j\in \mathcal{N}\left ( i \right )   \right \}  \right ) + Bias
\end{equation}

Here, we regard $f_{homo}$ as modeling solely based on the graph structure, and the error introduced by label autocorrelation is denoted by $Bias$. Thus, the residual of the graph-aware model is defined as:

\begin{equation}
    r_i = \hat{\mathcal{Y}}_i^{aware} - \hat{\mathcal{Y}}_i^{agnostic} = f_{homo}\left ( \left \{ x_j  \right \}  \right ) + Bias
\end{equation}

The bias is corrected by scaling the residual. $\beta_i$ has the same interpretation as the penalty factor in the negative feedback loss. The bias correction helps the predictions of the graph-aware model better approximate the ground truth $\mathcal{Y}$.

\begin{equation}
    \hat{\mathcal{Y}}_i^{correct} = \hat{\mathcal{Y}}_i^{aware} - \beta_i r_i
\end{equation}

For multi-layer GNN models, since bias propagates during each aggregation step, we apply negative feedback correction to the output of each layer. Specifically, the input to the $i$ layer of the model is obtained by applying negative feedback from the $i-1$ graph-agnostic layer. By correcting the output of each layer, the label autocorrelation bias can be minimized to the greatest extent.

\subsection{Dirichlet Energy-Based Feedback Coefficient}

In the previous sections, we introduce the bias correction mechanism based on the idea of negative feedback, which explicitly adjusts the prediction bias caused by label autocorrelation. Next, we will explain how to determine the appropriate value of the feedback factor. Since we aim to suppress label autocorrelation, the degree of homophily can be used to reflect the extent to which the bias needs to be corrected. Although the definition of label-based homophily is widely adopted, a significant limitation is that it may become ill-defined when labels are sparse in the underlying graph, which is often the case in practice. \cite{maskey2023fractional} uses Dirichlet energy to measure how much the features change in the nodes of $\mathcal{G}$ by quantifying the disparity between the normalized outflow of information from node $j$ and the normalized inflow of information to node $i$. 

\begin{equation}
    \mathscr{E}\left ( x \right ) = \frac{1}{4} \sum_{i,j=1}^{N}a_{i,j}\left \| \frac{x_i}{\sqrt{d^{in}_i} } - \frac{x_j}{\sqrt{d^{out}_j} } \right \| ^2_2 
\end{equation}

We extend it to undirected graphs and focus only on the feature representations of node $i$ and its neighboring nodes.

\begin{equation}
    \mathscr{E}\left ( i \right ) = \frac{1}{4}  \sum_{j\in \mathcal{N}\left ( i \right )  }^{} \left \| \frac{x_i}{\sqrt{\left | \mathcal{N}_i  \right | } } - \frac{x_j}{\sqrt{\left | \mathcal{N}_j  \right | } } \right \|^2_2 
\end{equation}

A smaller Dirichlet energy is regarded as indicating stronger homogeneity at the attribute level, and thus a larger feedback factor is required for bias correction. The entire learning procedure of the proposed GNFBC framework is summarized in Algorithm \ref{alg:algorithm}.

\begin{algorithm}[tb]
\caption{Optimization Procedure for GNFBC}
\label{alg:algorithm}
\textbf{Input}: Graph-aware model $\mathcal{M}_{aware}$, graph-agnostic counterpart $\mathcal{M}_{agnostic}$, graph $\mathcal{G} = \left \{ \mathcal{V},\mathcal{E},\mathcal{X},\mathcal{Y} \right \}$, number of training epochs $E$, learning rate $\alpha$, number of model layers $L$\\
\textbf{Initialize}: $\mathcal{M}_{aware}\left ( w \right )$ and $\mathcal{M}_{agnostic}\left ( w \right ) $ by Xavier\\
\textbf{Output}: Trained model parameters $w$\\
\begin{algorithmic}[1]
\FOR{$i=0;i< \left | \mathcal{V} \right |;i++$}
\STATE Calculate $\mathscr{E}\left ( i \right )$ by Eq.(17)
\STATE $\beta_i = 1-Norm\left ( \mathscr{E}\left ( i \right ) \right ) $
\ENDFOR
\FOR{$t=0;t<E;t++$}
\STATE $\hat{\mathcal{O}}^{correct} = \mathcal{X}$
\FOR{$l=0;i<L;l++$}
\STATE $\hat{\mathcal{O}}^{aware}= \mathcal{M}^{l}_{aware}\left ( \hat{\mathcal{O}}^{correct},\mathcal{E};w   \right ) $
\STATE $\hat{\mathcal{O}}^{agnostic}= \mathcal{M}^{l}_{agnostic}\left ( \hat{\mathcal{O}}^{correct};w   \right ) $
\STATE $\hat{\mathcal{O}}^{correct} = \left ( 1-\beta \right ) \hat{\mathcal{O}}^{aware} + \beta \hat{\mathcal{O}}^{agnostic}$
\ENDFOR
\STATE $\hat{\mathcal{Y}} = \hat{\mathcal{O}}^{correct}$
\STATE Calculate the total loss $L_{neg}$ by Eq.(10)
\STATE $w\gets w -  \alpha\bigtriangledown L_{neg}$
\ENDFOR
\STATE \textbf{return} $w$
\end{algorithmic}
\end{algorithm}

\subsection{Inference Process}

By introducing a graph-agnostic counterpart with shared parameters during training, the proposed negative feedback correction effectively mitigates the over-reliance on label autocorrelation while incurring no additional inference cost. This design allows the model to preserve the computational and deployment efficiency of a standard graph-aware model at inference time, while benefiting from improved generalization and robustness to heterogeneity and structural noise, as the correction effect is already embedded into the shared parameters learned during training. We will further analyze the time complexity of GNFBC in the experimental section.

\section{Experiments}

GraphSAGE \cite{GraphSAGE} exhibits strong performance and can effectively scale to large-scale graphs. Therefore, in the main experiments, we selected GraphSAGE as the graph-aware model and used a parameter-sharing MLP as the graph-agnostic model for negative feedback. Furthermore, we conduct experiments with other GNNs to thoroughly validate the effectiveness and robustness of the GNFBC framework.

\subsection{Datasets}

We evaluate our methods on datasets with varying degrees of heterophily. Specifically, we utilize five homophilic graphs, including Cora, CiteSeer, PubMed \cite{PubMed}, Computers, and Photo, which are widely used citation and co-purchase networks. To assess performance under heterohpily, we also consider four benchmark heterophilic graphs: Wisconsin, Washington, Texas, and Cornell, which are derived from the WebKB collection. In addition, we also use the YelpChi \cite{YelpChi} and Amazon \cite{Amazon} datasets. These two datasets exhibit a moderate level of heterogeneity, with the heterogeneous and homogeneous components within the graph being roughly balanced. Detailed information on these datasets can be found in \textbf{Appendix A}.

\subsection{Experimental Setup}

\subsubsection{Comparison Methods}

We employ several different groups of baseline methodologies for comparison with the proposed framework. 1) Non-GNN methods: RF \cite{RF}, MLP \cite{MLP}. 2) Homophilic GNNs: SGC \cite{SGC}, GCN \cite{GCN}, GAT \cite{GAT}. 3) Heterophilic GNNs: GPRGNN \cite{GPRGNN}, H\textsubscript{2}GCN \cite{H2GCN}, FAGCN \cite{FAGCN}, RAW-GNN \cite{RAW-GNN}, RUM \cite{RUM}, LG-GNN \cite{LG-GNN}. For the commonly used YelpChi and Amazon datasets in fraud detection research, we additionally select task-specific baselines CARE-GNN \cite{CARE-GNN}, PC-GNN \cite{PC-GNN}, RioGNN \cite{RioGNN}, H2-FDetector \cite{H2-FDetector}, GTAN \cite{GTAN} and ConsisGAD \cite{ConsisGAD}.

\subsubsection{Implementation Details}

The implementation of GNFBC is based on PyTorch, and all experiments are conducted on an NVIDIA RTX 4090 GPU. We adopt a data split ratio
of 40\% for training, 20\% for validation, and 40\% for the test set. The model parameters are initialized using the Xavier initialization method and optimized with Adam. We adopt early stopping and terminating training if the validation performance does not improve for 20 consecutive epochs.

\subsubsection{Metrics}

For evaluation, we primarily report classification accuracy, which reflects the proportion of correctly classified nodes and is the standard metric for node classification tasks. For YelpChi and Amazon, which are binary classification tasks with highly imbalanced label distributions, we additionally report the Area Under the ROC Curve (AUC) and the F1-Macro, as they better capture the discriminative ability of the model and the balance between precision and recall under class imbalance.

\begin{table*}[ht]
\centering
\begin{tabular}{l|ccccc|cccc}
\toprule
 & \multicolumn{5}{c|}{\textbf{Homophilic Graphs}} & \multicolumn{4}{c}{\textbf{Heterophilic Graphs}} \\
\cmidrule(lr){2-6} \cmidrule(lr){7-10}
 & Cora & CiteSeer & PubMed & Computers & Photo & Chameleon & Squirrel & Texas & Cornell \\
\midrule
RF        & 79.53 & 68.28 & 84.56 & 73.42 & 77.25 & 53.92 & 38.76 & 85.07 & 82.48 \\
MLP     & 71.39 & 66.58 & 82.29 & 70.46 & 78.61 & 46.27 & 31.55 & 81.63 & 83.92 \\
\midrule
SGC        & 69.74 & 58.34 & 81.56 & 75.28 & 81.88 & 64.57 & 43.17 & 55.26 & 46.83 \\
GCN       & 75.18 & 67.89 & 85.43 & 81.89 & 89.32 & 60.78 & 45.33 & 55.03 & 61.82 \\
GAT       & 76.89 & 67.56 & 83.46 & 82.13 & 90.32 & 63.46 & 43.11 & 58.98 & 60.03 \\
\midrule
GPRGNN  & 77.93 & 67.33 & 84.09 & \textbf{81.37} & 90.52 & 65.43 & 49.89 & 82.59 & 84.74 \\
H\textsubscript{2}GCN & 83.35 & 72.89 & \underline{88.47} & 80.13 & 86.59 & 62.39 & 48.63 & 80.95 & 80.72 \\
FAGCN      & 84.01 & 71.98 & 79.63 & 79.73 & 88.54 & \underline{66.75} & \underline{57.46} & 84.38 & 79.58 \\
RAW-GNN & \underline{85.32} & 70.49 & 83.62 & 78.86 & 89.14 & 64.93 & 51.28 & 83.48 & \underline{84.77} \\
RUM      & 82.13 & 72.64 & 82.22 & 77.83 & \underline{91.52} & 62.78 & 52.33 & 80.07 & 71.28 \\
LG-GNN & 84.78 & \underline{73.34} & 88.42 & 76.42 & \textbf{91.73} & 66.21 & 56.74 & \underline{88.36} & 81.43 \\
\midrule
\textbf{GNFBC} & \textbf{86.56} & \textbf{73.91} & \textbf{89.30} & \underline{80.43} & 89.54 & \textbf{67.42} & \textbf{59.31} & \textbf{89.90} & \textbf{88.89} \\

GNFBC(w/o $L_{neg}$) & 86.25 & 72.97 & 89.12 & 78.94 & 88.76 & 66.03 & 57.54 & 88.21 & 86.48 \\

\bottomrule
\end{tabular}
\caption{Performance comparison on five homophilic and four heterophilic graph datasets. The best result is shown in \textbf{bold}, while the second best is marked with \underline{underline}.}
\label{tab:main_exp_1}
\end{table*}

\subsection{Performance Evaluation}

As shown in Table \ref{tab:main_exp_1}, we conduct experiments on five homophilic and four heterophilic graph datasets to compare the performance of different methods for node classification. GNFBC performs best on 7 out of 9 datasets. The following is a detailed explanation.
Compared to traditional GNNs, GNFBC achieves an average performance improvement of 7.92\% to 36.92\% in all datasets, with the highest gains observed in heterophilic datasets, particularly Texas and Cornell. Against heterophilic GNNs, GNFBC shows an average improvement of 3.56\%, with the most significant gains in Cornell and Squirrel. The results highlight GNFBC's ability to adaptively model homophilic and heterophilic graph structures, particularly excelling in heterophilic settings where traditional GNNs falter due to their reliance on mean aggregation, which fails to capture dissimilar neighbor patterns.

The experimental results in Table \ref{tab:main_exp_2} further demonstrate the robustness of GNFBC in various heterogeneity scenarios. On YelpChi, GNFBC outperforms traditional GNNs by up to 33.07\%. Against heterophilic GNNs, it achieves gains of up to 14.14\%. GNFBC also exhibits comparable performance on the Amazon dataset. The Yelpchi and Amazon datasets contain both local homogeneity and local heterogeneity. Our proposed framework can effectively adapt to this task by adjusting the degree of feedback correction for each node.

\begin{table}[ht]
\centering
\begin{tabular}{l|cccc}
\toprule
\multirow{2}{*}{} & \multicolumn{2}{c}{\textbf{YelpChi}} & \multicolumn{2}{c}{\textbf{Amazon}} \\
\cmidrule(lr){2-3} \cmidrule(lr){4-5}
 & AUC & F1 & AUC & F1 \\
\midrule
RF     & 86.49 & 70.24 & 94.33 & 91.50 \\
MLP      & 69.25 & 54.86 & 92.76 & 90.02 \\
\midrule
SGC        & 61.27 & 48.58 & 80.13 & 76.59 \\
GCN     & 61.03 & 48.32 & 80.49 & 76.93 \\
GAT      & 60.39 & 47.53 & 83.87 & 79.20 \\
\midrule
CARE-GNN         & 77.20 & 61.16 & 90.41 & 86.44 \\
PC-GNN         & 81.76 & 68.39 & 96.65 & 88.17 \\
RioGNN          & 81.85 & 65.40 & 97.29 & 88.08 \\
\midrule
H2-FDetector  & \underline{89.62} & 71.47 & 96.82 & 80.08 \\
GTAN          & 90.38 & 77.33 & 93.42 & 91.03 \\
ConsisGAD     & 89.16 & \underline{75.14} & \underline{97.15} & \textbf{92.76} \\
\midrule
\textbf{GNFBC} & \textbf{91.34} & \textbf{78.54} & \textbf{97.55} & \underline{92.49} \\
\bottomrule
\end{tabular}
\caption{Performance comparison on YelpChi and Amazon datasets in terms of AUC and F1-macro (\%).}
\label{tab:main_exp_2}
\end{table}

\subsection{Ablation Study}

\begin{figure}[htbp]
    \centering
    \includegraphics[width=0.98\linewidth]{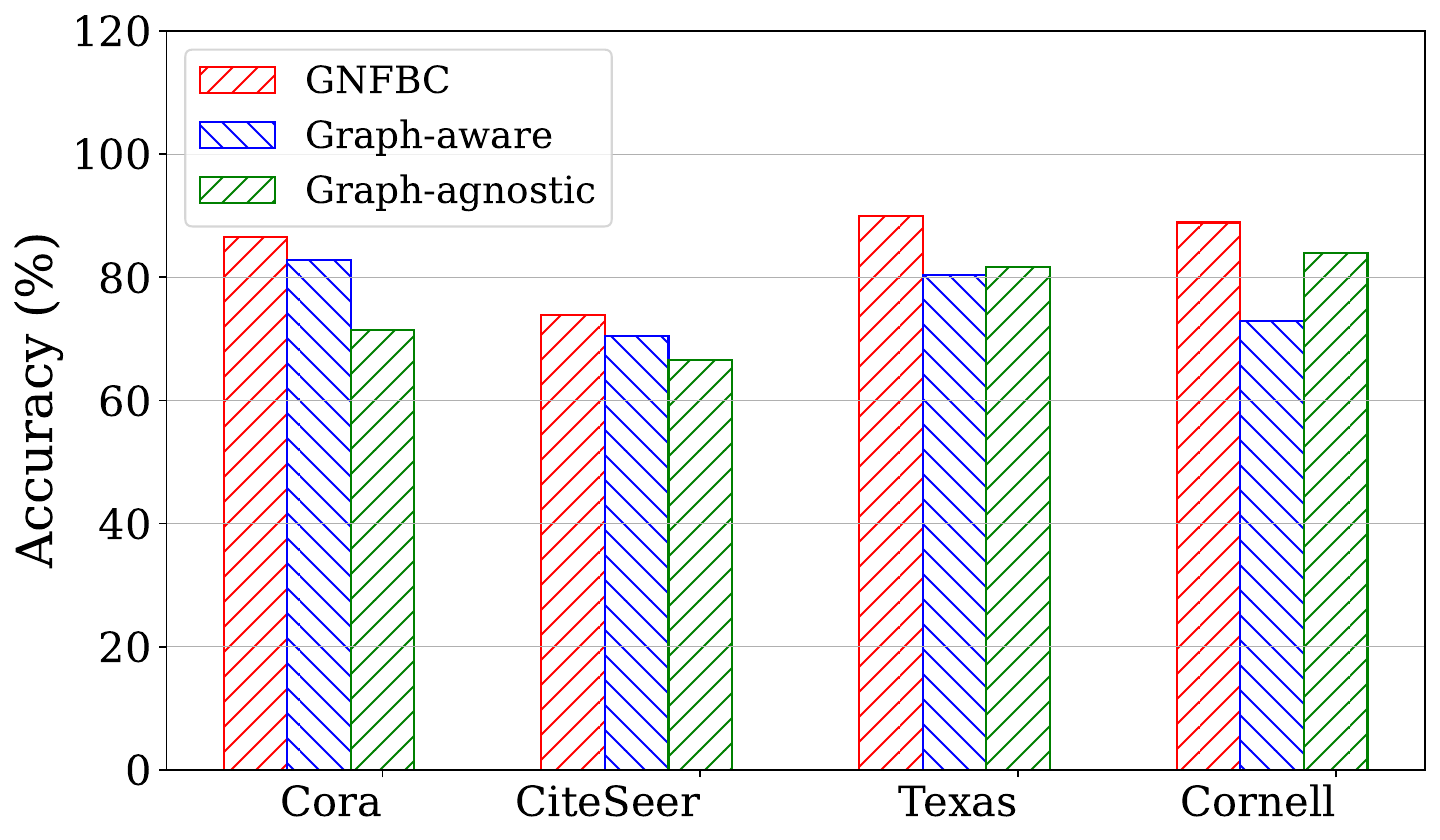}
    \caption{Performance comparison between GNFBC and standalone Graph-aware or Graph-agnostic model on homophilic and heterophilic datasets.}
    \label{fig:ablation_homo_hete}
\end{figure}

\begin{table}[h]
\centering
\begin{tabular}{l|ccc}
\toprule
 & GNFBC & Graph-aware & Graph-agnostic \\
\midrule
YelpChi & \textbf{91.34} & 80.87 & 69.25 \\
Amazon  & \textbf{97.55} & 94.01 & 92.76 \\
\bottomrule
\end{tabular}
\caption{Performance comparison between GNFBC and standalone Graph-aware or Graph-agnostic model on YelpChi and Amazon datasets.}
\label{tab:ablation_yelpchi_amazon}
\end{table}

\begin{figure*}[htbp]
    \centering
    \includegraphics[width=0.32\linewidth]{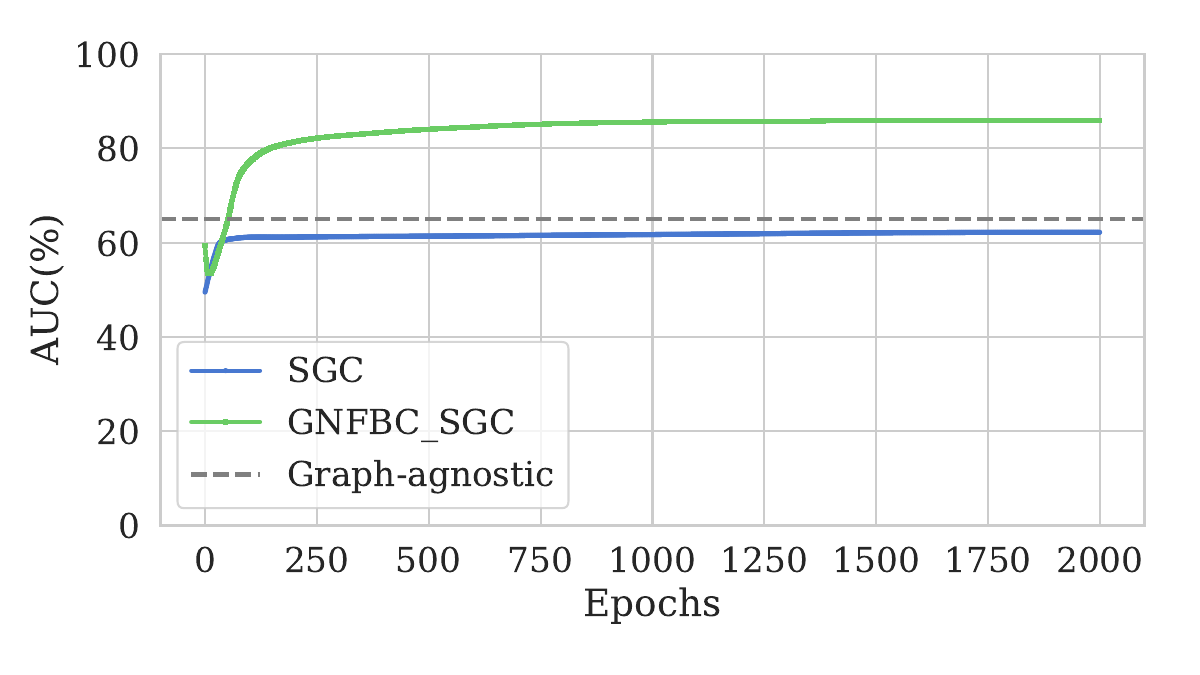}
    \hfill
    \includegraphics[width=0.32\linewidth]{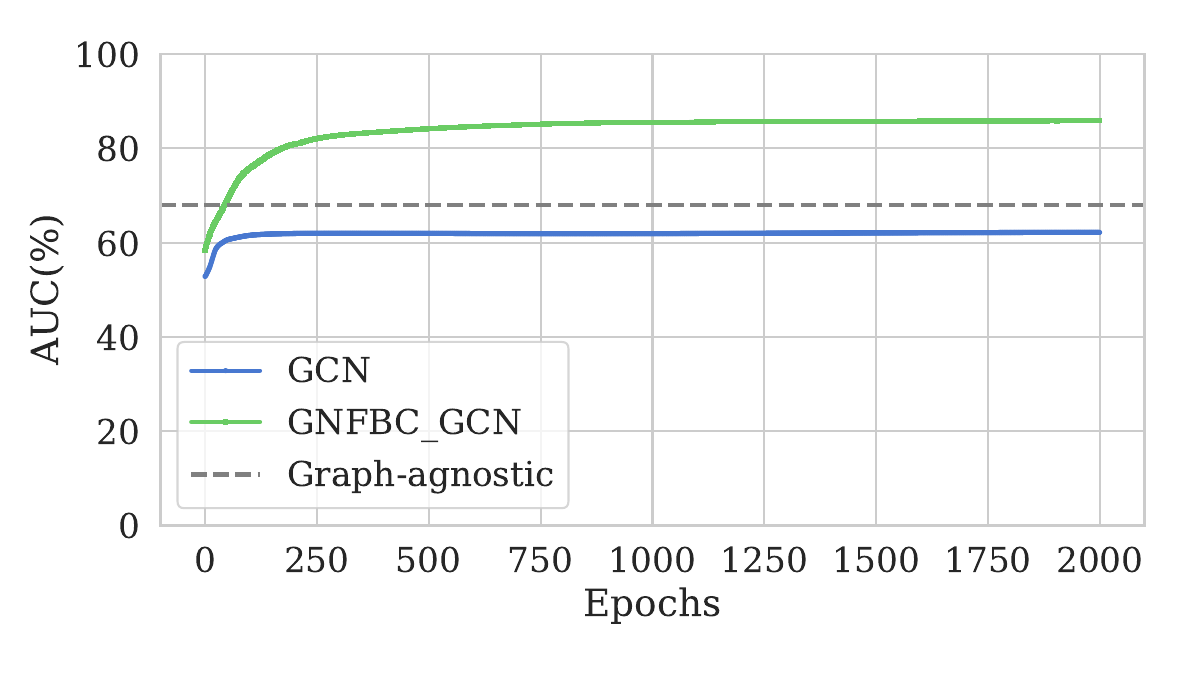}
    \hfill
    \includegraphics[width=0.32\linewidth]{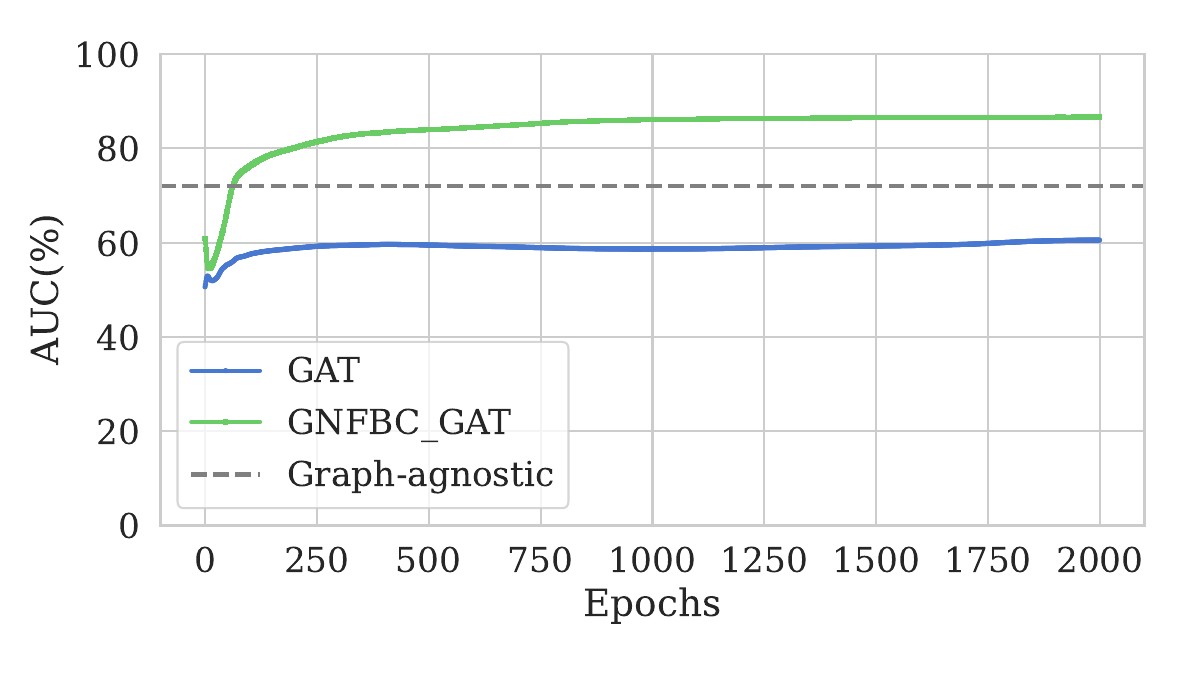}
    \caption{Performance of SGC, GCN, and GAT models with the GNFBC framework on the YelpChi dataset}
    \label{fig:robust_yelpchi}
\end{figure*}

\begin{figure*}[htbp]
    \centering
    \includegraphics[width=0.32\linewidth]{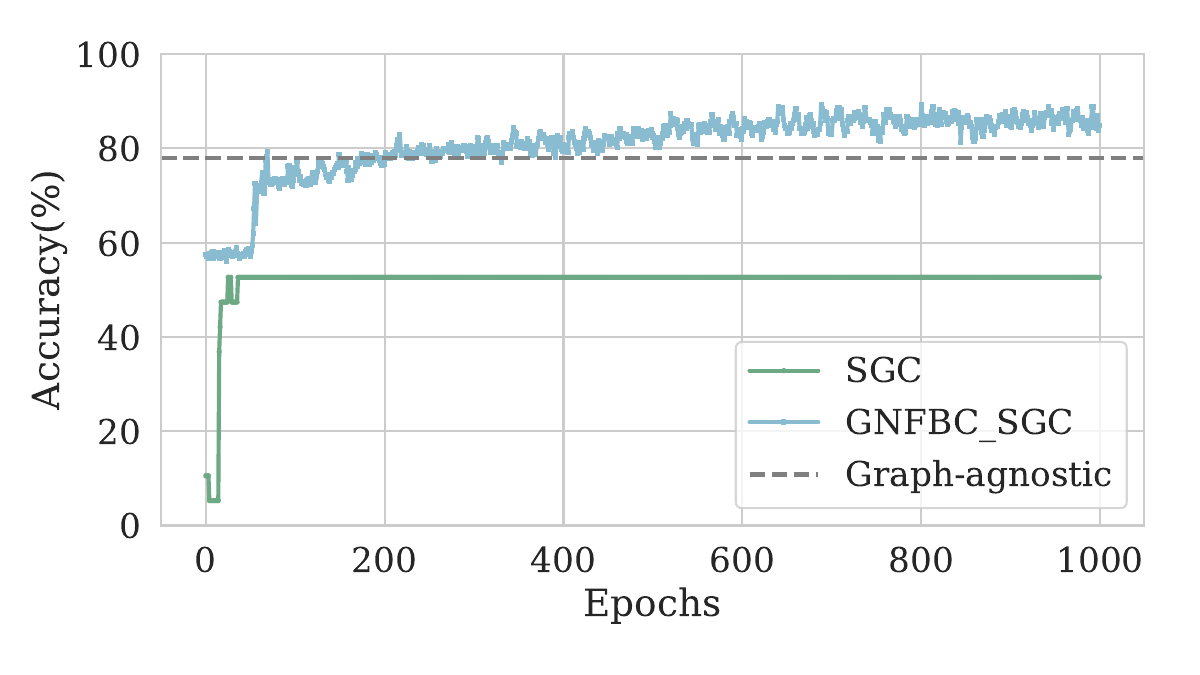}
    \hfill
    \includegraphics[width=0.32\linewidth]{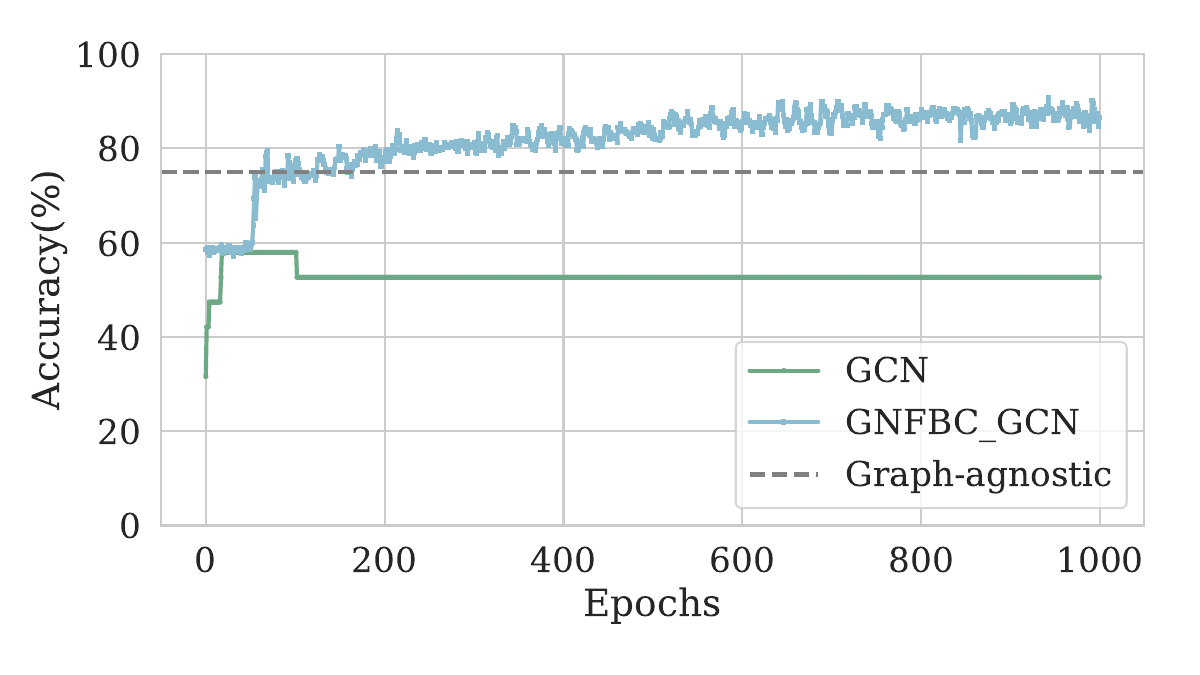}
    \hfill
    \includegraphics[width=0.32\linewidth]{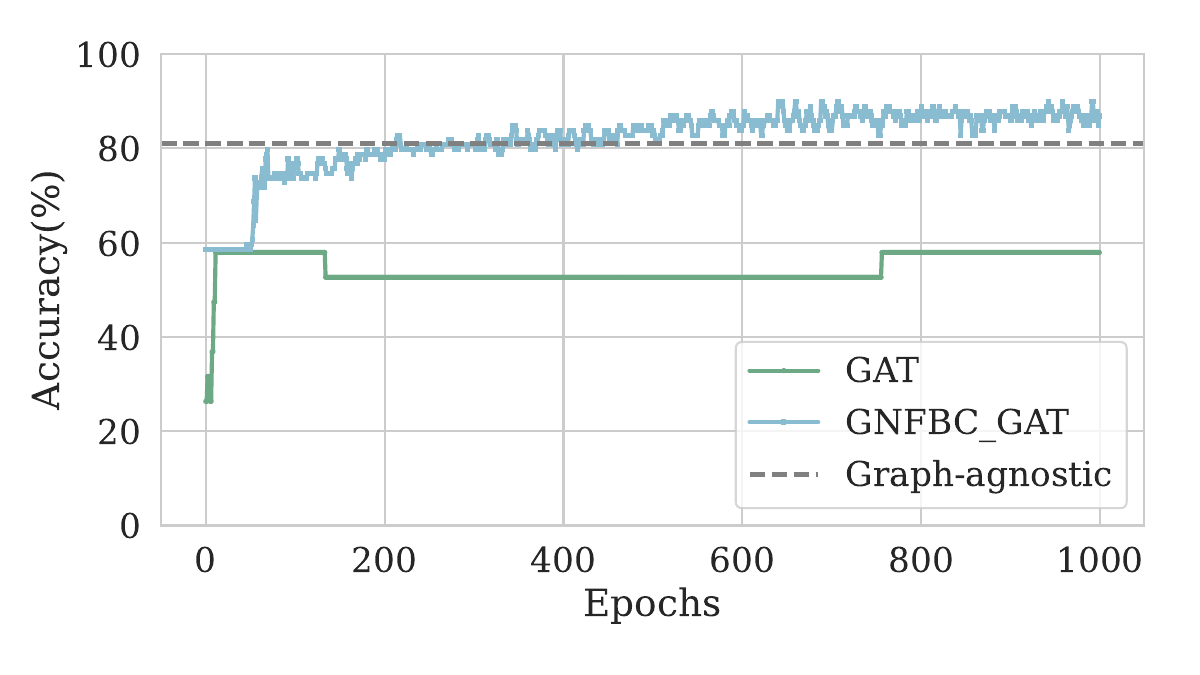}
    \caption{Performance of SGC, GCN, and GAT models with the GNFBC framework on the Texas dataset}
    \label{fig:robust_texas}
\end{figure*}

GNFBC mitigates the label autocorrelation problem in GNN models by introducing a negative feedback loss and a graph-agnostic model feedback correction mechanism. To verify the importance of these two components, we conduct ablation studies. Specifically, we remove negative feedback loss from the GNFBC and conduct experiments again. As shown in Table \ref{tab:main_exp_1}, performance degrades to varying degrees in all datasets, with a more pronounced effect on heterogeneous graphs. This is because label autocorrelation has a greater impact on heterogeneous graphs.

We further analyze the performance improvement of GNFBC compared to using the graph-aware model or the graph-agnostic model alone. As shown in Figure \ref{fig:ablation_homo_hete}, GNFBC achieves better performance in both homophilic datasets (Cora, CiteSeer) and heterophilic datasets (Texas, Cornell). Meanwhile, since the graph-aware model relies on the homophily assumption, it performs better on homophilic datasets. In contrast, the graph-agnostic model performs relatively better on heterophilic datasets. It should be noted that GNFBC also achieves significant improvements in the YelpChi and Amazon datasets. As shown in Table \ref{tab:ablation_yelpchi_amazon}, 
GNFBC improves the AUC by 10.47\% on YelpChi and by 3.54\% on Amazon compared to the graph-aware model. This further demonstrates the robustness of GNFBC across datasets with varying degrees of heterogeneity.

\subsection{Robustness Research}

Next, we conduct experiments on several other classic GNN models. Specifically, we treat SGC, GCN, and GAT as graph-aware models, and apply their corresponding graph-agnostic models for negative feedback bias correction. As shown in Figure \ref{fig:robust_yelpchi}, on the YelpChi dataset, the three homophilic GNNs achieve significant improvements under the GNFBC framework, with an average increase in AUC of 24.28\%. Similarly, on the heterogeneous Texas dataset, GNFBC also demonstrates strong performance across different GNN models. As shown in Figure \ref{fig:robust_texas}, the accuracy improves by almost 30\% compared to the original models.

\subsection{Computational and Memory Overhead}

We analyze the computational and memory overhead introduced by incorporating the negative feedback correction mechanism into GraphSAGE. For a mini-batch of size $B$, GraphSAGE adopts neighbor sampling at each of the $L$ layers, with a fixed fan-out of $S_l$ at layer $l$. The total computational complexity of GraphSAGE is approximately as follows.
\begin{equation}
    \mathcal{O}\left ( B\sum_{l=1}^{L}\left ( \prod_{i=1}^{l} S_i  \right )  d_lh_l \right ) 
\end{equation}

In the proposed framework, we simultaneously run a graph-agnostic model to compute a residual signal for bias correction. We enforce parameter sharing between the graph-aware and graph-agnostic models to avoid additional parameter storage. Since the graph-agnostic model only performs fully connected transformations without neighborhood aggregation, its computational complexity is $\mathcal{O}\left ( BLd_lh_l \right ) $, which is negligible compared to GraphSAGE. Similarly, the additional memory footprint comes mainly from storing intermediate activations of the graph-agnostic model, which scales linearly with the batch size and number of layers and is significantly smaller than the memory required for the neighborhood-expanded activations of GraphSAGE.

Therefore, the negative feedback correction mechanism introduces only marginal computational and memory overhead while effectively improving prediction performance by mitigating structural bias.

\section{Conclusion}

This paper addresses biases in graph neural networks (GNNs) caused by label autocorrelation under the homophily assumption. We propose the Graph Negative Feedback Bias Correction (GNFBC) framework, which mitigates these biases using a negative feedback loss and graph-agnostic model feedback correction, guided by Dirichlet energy. GNFBC can be seamlessly integrated into almost any GNN architecture, improving overall model performance with comparable computational and memory overhead.

\bibliography{aaai2026}

\clearpage

\appendix

\section{A. Dataset Description}

We evaluate our methods on a diverse set of graph datasets, categorized based on their homophily and heterophily characteristics. Below, we provide a detailed description of each dataset, followed by a summary table outlining key statistics, including the number of nodes, edges, and other relevant metrics.
\subsection{Homophilic Datasets}
These datasets exhibit strong homophily, where connected nodes tend to share similar labels or attributes. They are commonly used in graph-based learning tasks:
\begin{itemize}
    \item \textbf{Cora}: A citation network dataset where nodes represent machine learning papers, and edges denote citation relationships. Each node is labeled with one of seven research topics.
    \item \textbf{CiteSeer}: Another citation network, similar to Cora, where nodes are scientific publications, and edges represent citations. Labels correspond to six research fields.
    \item \textbf{PubMed}: A citation network of biomedical research papers, with nodes labeled according to diabetes-related categories (three classes). Edges represent citations between papers.
    \item \textbf{Computers}: A co-purchase network from Amazon, where nodes represent computer products, and edges indicate frequent co-purchases. Nodes are labeled by product category.
    \item \textbf{Photo}: Another Amazon co-purchase network, focusing on photography-related products. Nodes are labeled by product category, with edges indicating co-purchase relationships.
\end{itemize}

\subsection{Heterophilic Datasets}
These datasets are characterized by heterophily, where connected nodes often have different labels or attributes. They are derived from the WebKB collection:
\begin{itemize}
    \item \textbf{Wisconsin}: A webpage network from the University of Wisconsin, where nodes are webpages, and edges represent hyperlinks. Nodes are labeled by webpage type (e.g., faculty, student).
    \item \textbf{Washington}: Similar to Wisconsin, this dataset contains webpages from the University of Washington, with hyperlinks as edges and webpage types as labels.
    \item \textbf{Texas}: A webpage network from the University of Texas, structured similarly to Wisconsin and Washington, with nodes labeled by webpage category.
    \item \textbf{Cornell}: A webpage network from Cornell University, following the same structure as the other WebKB datasets, with nodes representing webpages and edges denoting hyperlinks.
\end{itemize}

\subsection{Moderately Heterogeneous Datasets}
These datasets exhibit a balanced mix of homophilic and heterophilic connections:
\begin{itemize}
    \item \textbf{YelpChi}: A dataset derived from Yelp reviews, where nodes represent businesses or users, and edges capture interactions (e.g., reviews or ratings). The dataset has a moderate level of heterogeneity, with a mix of homophilic and heterophilic connections.
    \item \textbf{Amazon}: A dataset from Amazon, representing a product co-purchase or user interaction network. It exhibits moderate heterogeneity, with balanced homophilic and heterophilic components.
\end{itemize}

\subsection{Dataset Statistics}
The following table summarizes key statistics for each dataset, including the number of nodes, edges, and classes.

\begin{table}[h]
\centering
\begin{tabular}{lccc}
\toprule
\textbf{Dataset} & \textbf{Nodes} & \textbf{Edges} & \textbf{Classes} \\
\midrule
Cora & 2,708 & 5,429 & 7 \\
CiteSeer & 3,327 & 4,732 & 6 \\
PubMed & 19,717 & 44,338 & 3 \\
Computers & 13,752 & 245,861 & 10 \\
Photo & 7,650 & 119,081 & 8 \\
Wisconsin & 251 & 499 & 5 \\
Washington & 230 & 446 & 5 \\
Texas & 183 & 309 & 5 \\
Cornell & 183 & 295 & 5 \\
YelpChi & 45,954 & 3,846,979 & 2 \\
Amazon & 11,944 & 4,398,392 & 2 \\
\bottomrule
\end{tabular}
\caption{Summary of Dataset Statistics}
\end{table}

\section{B. Theoretical Analysis of Bias in Graph Attention Networks}

The autocorrelation of neighbor labels is a key characteristic of graph data, influencing the performance of Graph Neural Networks (GNNs). Graph Attention Networks (GATs) utilize an attention-based aggregation strategy to dynamically weigh neighbor contributions. This section analyzes how label autocorrelation introduces bias in GATs for node classification, integrating the aggregation strategy into the bias formula for further derivation.
\subsection{GAT Aggregation Strategy}

For a graph $\mathcal{G} = (\mathcal{V}, \mathcal{E})$ with node features $\mathcal{X} \in \mathbb{R}^{N \times F}$, GATs compute the representation of node $i$ at layer $l+1$ as:

\begin{equation}
    \mathbf{h}_i^{(l+1)} = \sigma \left( \sum_{j \in \mathcal{N}(i) \cup \{i\}} \alpha_{ij}^{(l)} \mathbf{W}^{(l)} \mathbf{h}_j^{(l)} \right),
\end{equation}

where $\mathbf{h}_i^{(l)}$ is the feature vector of node $i$ at layer $l$, $\mathbf{W}^{(l)}$ is a learnable weight matrix, $\sigma$ is a non-linear activation (e.g., ReLU), and $\alpha_{ij}^{(l)}$ is the attention coefficient, defined as:

\begin{equation}
    \alpha_{ij}^{(l)} = \frac{\exp \left( a(\mathbf{h}_i^{(l)}, \mathbf{h}_j^{(l)}) \right)}{\sum_{k \in \mathcal{N}(i) \cup \{i\}} \exp \left( a(\mathbf{h}_i^{(l)}, \mathbf{h}_k^{(l)}) \right)},
\end{equation}

with $a(\cdot, \cdot)$ being a scoring function (e.g., based on feature concatenation and a learnable vector). For simplicity, we model the aggregation as a weighted sum of neighbor features, where $\alpha_{ij}^{(l)}$ reflects the importance of neighbor $j$ to node $i$. The final node prediction $\hat{\mathcal{Y}}_i$ is derived from the output layer, typically via a softmax function.
\subsection{Bias Derivation with GAT Aggregation}

To quantify the bias due to label autocorrelation, we focus on first-order neighbor dependencies. The prediction error for node $i$, incorporating neighbor influences, is:

\begin{equation}
    \epsilon_i = \hat{\mathcal{Y}}_i + \sum_{j \in \mathcal{N}(i)} \rho_{ij} (\mathcal{Y}_j - \hat{\mathcal{Y}}_j),
\end{equation}

where $\hat{\mathcal{Y}}_i$ is the predicted label, $\mathcal{Y}_j$ is the true label of neighbor $j$, and $\rho_{ij}$ is the partial correlation between labels $\mathcal{Y}_i$ and $\mathcal{Y}_j$. In GATs, the prediction $\hat{\mathcal{Y}}_i$ depends on the aggregated representation:

\begin{equation}
    \hat{\mathcal{Y}}_i = f \left( \sum_{j \in \mathcal{N}(i) \cup \{i\}} \alpha_{ij} \mathbf{W} \mathbf{h}_j \right),
\end{equation}

where $f$ is the output function (e.g., softmax), and $\alpha_{ij}$ weights the contribution of neighbor $j$. We approximate $\rho_{ij}$ as being influenced by $\alpha_{ij}$, since the attention mechanism emphasizes neighbors with similar features, which often correlate with similar labels. Thus, we model $\rho_{ij} \approx \kappa \alpha_{ij}$, where $\kappa$ is a scaling factor reflecting the correlation strength.

Substituting into the error term:

\begin{equation}
    \epsilon_i = \hat{\mathcal{Y}}_i + \kappa \sum_{j \in \mathcal{N}(i)} \alpha_{ij} (\mathcal{Y}_j - \hat{\mathcal{Y}}_j).
\end{equation}

The bias in the learning objective, compared to the negative log-likelihood (NLL) of the true data, is:

\begin{equation}
    \text{Bias} \approx \sum_{i=1}^N \frac{1}{2 \sigma^2} (\mathcal{Y}_i - \hat{\mathcal{Y}}_i)^2 - \sum_{i=1}^N \frac{1}{2 \sigma^2 (1 - \rho_i^2)} \left( \mathcal{Y}_i - \epsilon_i \right)^2,
\end{equation}

where $\rho_i^2 = \sum_{j \in \mathcal{N}(i)} \rho_{ij}^2 \approx \kappa^2 \sum_{j \in \mathcal{N}(i)} \alpha_{ij}^2$, and $\sigma^2$ is the label distribution variance. The second term reflects the adjusted error, accounting for the attention-weighted influence of neighbor labels. This expression shows that the attention coefficients $\alpha_{ij}$ directly influence the bias through their role in weighting neighbor contributions. High $\alpha_{ij}$ values for neighbors with correlated labels (i.e., large $\mathcal{Y}_j - \hat{\mathcal{Y}}_j$) amplify the bias, as the model overestimates the graph's explanatory power.

\end{document}